%% file: main_arxiv.tex
\documentclass[10pt,twocolumn,letterpaper]{article}

\usepackage[pagenumbers]{cvpr} 


\definecolor{cvprblue}{rgb}{0.21,0.49,0.74}
\usepackage[pagebackref,breaklinks,colorlinks,allcolors=cvprblue]{hyperref}
\usepackage{multirow}
\usepackage{colortbl}

\def\paperID{8689} 
\def\confName{CVPR}
\def\confYear{2025}

\title{EDCFlow: Exploring Temporally Dense Difference Maps for Event-based Optical Flow Estimation}

\author{{Daikun Liu}\qquad{Lei Cheng}\qquad{Teng Wang\thanks{Corresponding author:~Teng Wang.}}\qquad{Changyin Sun}\\
Southeast University\\
{\tt\small \{dkliu, leicheng, wangteng, cysun\}@seu.edu.cn}}

\begin{document}
\maketitle
\input{sec/0_abstract}    
\input{sec/1_intro}
\input{sec/2_related_work}
\input{sec/3_method}

\input{sec/4_exper_results}
\input{sec/5_conclusion}

{
    \small
    \bibliographystyle{ieeenat_fullname}
    \bibliography{main}
}

\input{sec/X_suppl}

\end{document}

%% file: sec/0_abstract.tex
\begin{abstract}
Recent learning-based methods for event-based optical flow estimation utilize cost volumes for pixel matching but suffer from redundant computations and limited scalability to higher resolutions for flow refinement. In this work, we take advantage of the complementarity between temporally dense feature differences of adjacent event frames and cost volume and present a lightweight event-based optical flow network (EDCFlow) to achieve high-quality flow estimation at a higher resolution. Specifically, an attention-based multi-scale temporal feature difference layer is developed to capture diverse motion patterns at high resolution in a computation-efficient manner. An adaptive fusion of high-resolution difference motion features and low-resolution correlation motion features is performed to enhance motion representation and model generalization. Notably, EDCFlow can serve as a plug-and-play refinement module for RAFT-like event-based methods to enhance flow details. Extensive experiments demonstrate that EDCFlow achieves better performance with lower complexity compared to existing methods, offering superior generalization. 



\end{abstract}

%% file: sec/1_intro.tex
\section{Introduction}
\label{sec:intro}

Optical flow estimation calculates per-pixel displacement vectors and is important for various applications like deblurring~\cite{debluring1, debluring2}, action recognition~\cite{action1, action2}, and object tracking~\cite{tracking1}. Event streams offer fine-grained temporal detail, high dynamic range, and no motion blur~\cite{survey}, making them well-suited for motion capture in challenging scenarios.

\begin{figure}[t]
\centering
\includegraphics[width=0.85\linewidth, height=0.28\textwidth]{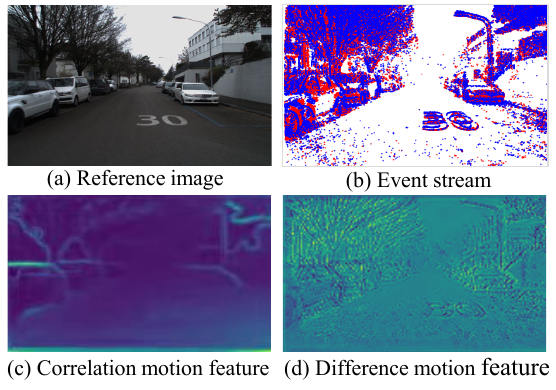}
\caption{Illustration of difference between correlation and difference motion features. (a) Reference image of DSEC~\cite{dsec} dataset. (b) Corresponding event stream. (c) Correlation motion feature. (d) Difference motion feature.}
\label{fig_1}
\end{figure}

Flow estimation essentially involves establishing correspondences between the source and target pixels, commonly achieved by modeling correlations in 4D cost volumes~\cite{raft, flownet1, pwc}. As a representative method, RAFT~\cite{raft} constructs a single-resolution (\textit{i.e.}, 1/8) cost volume of all-pairs for correlations retrieval and employs a GRU for iterative refinement. Inspired by the success of RAFT in the frame domain, several event-based methods~\cite{DCEIFlow, e-raft, TMA, multiflow} adopt analogous architectures by treating event streams as consecutive event frames. Specifically, DCEIFlow~\cite{DCEIFlow} and E-RAFT~\cite{e-raft} focus on long-range correspondences, neglecting intermediate motions in events; TMA~\cite{TMA} and MultiFlow~\cite{multiflow} mitigate this by building temporally dense cost volumes, resulting in a complexity of $\mathcal{O}(TN^{2}C)$ with $T$, $N$, and $C$ representing the number of event frames, pixels, and feature dimensions, respectively. However, event streams typically provide small displacement in a short time window and performing global 2D searches within temporally dense cost volumes leads to information redundancy and unnecessary computations. Furthermore, this computational burden limits the scalability of these RAFT-like methods to higher resolutions, which offer potential accuracy gains as evidenced by recent studies~\cite{emd, idnet}.

In this work, we attempt to perform flow estimation at a higher resolution (\textit{i.e.}, 1/4) without resorting to computationally expensive high-resolution cost volumes. Instead, we exploit the potential of temporally dense motion features derived from dense feature differences between adjacent event frames. As illustrated in~\cref{fig_1}, cost volumes reflect global similarity and are robust to noise, but prone to matching ambiguities and expensive computations. In contrast, feature differences capture details with clear motion boundaries and are computationally efficient (\textit{i.e.}, $\mathcal{O}(TNC)$), but are sensitive to noise, limiting their Generalizability. This complementarity motivates us to present a joint difference and correlation network for event-based optical flow, dubbed \textbf{EDCFlow}, as shown in Fig.~\ref{fig_2}, that iteratively integrates motion features from temporally dense high-resolution (\textit{i.e.}, 1/4) feature difference maps and a low-resolution (\textit{i.e.}, 1/8) cost volume to refine flows, thereby achieving better accuracy with lower complexity. 

To be specific, in each iteration, we look up the correlation, upsample it to the higher resolution, and encode correlation motion features. To capture intermediate motions, we discretize moving trajectories and compute the feature differences between adjacent frames. More precisely, we perform feature warping on consecutive frames, and introduce a multi-scale temporal feature difference layer with attention to capture various motion patterns, providing reliable motion details for both large and small displacements. The correlation and difference motion features are finally adaptively fused to update the flow. Notably, our model can be cascaded into existing RAFT-like networks to enhance flow quality, especially at motion boundaries, with marginal additional parameters and computations. Extensive experimental results on DSEC~\cite{dsec} and MVSEC~\cite{mvsec} demonstrate that EDCFlow achieves SOTA-comparable performance with higher efficiency, offering advantages for real-world deployment and adoption. Furthermore, we conduct ablation studies to validate key design choices of EDCFlow. We summarize our contributions as follows:

\begin{itemize}
         \item We introduce a lightweight event-based optical flow network, which explores the temporally dense feature differences between adjacent event frames, combined with a low-resolution cost volume to enable high-quality flow estimation at a higher resolution. 
         \item We construct an attention-based feature difference layer to aggregate temporally multi-scale motion patterns and adaptively fuse complementary motion features from difference maps and cost volume, enhancing both representation power and generalization capability of our model.
         \item The proposed methods can be flexibly integrated into event-based RAFT-like networks to further refine flow.
	\item Our method achieves state-of-the-art or comparable performance on DSEC and MVSEC while striking a better balance between accuracy, model size, and computations. 
\end{itemize}

%% file: sec/2_related_work.tex
\begin{figure*}[t]
\centering
\includegraphics[width=0.9\textwidth,height=0.38\textwidth]{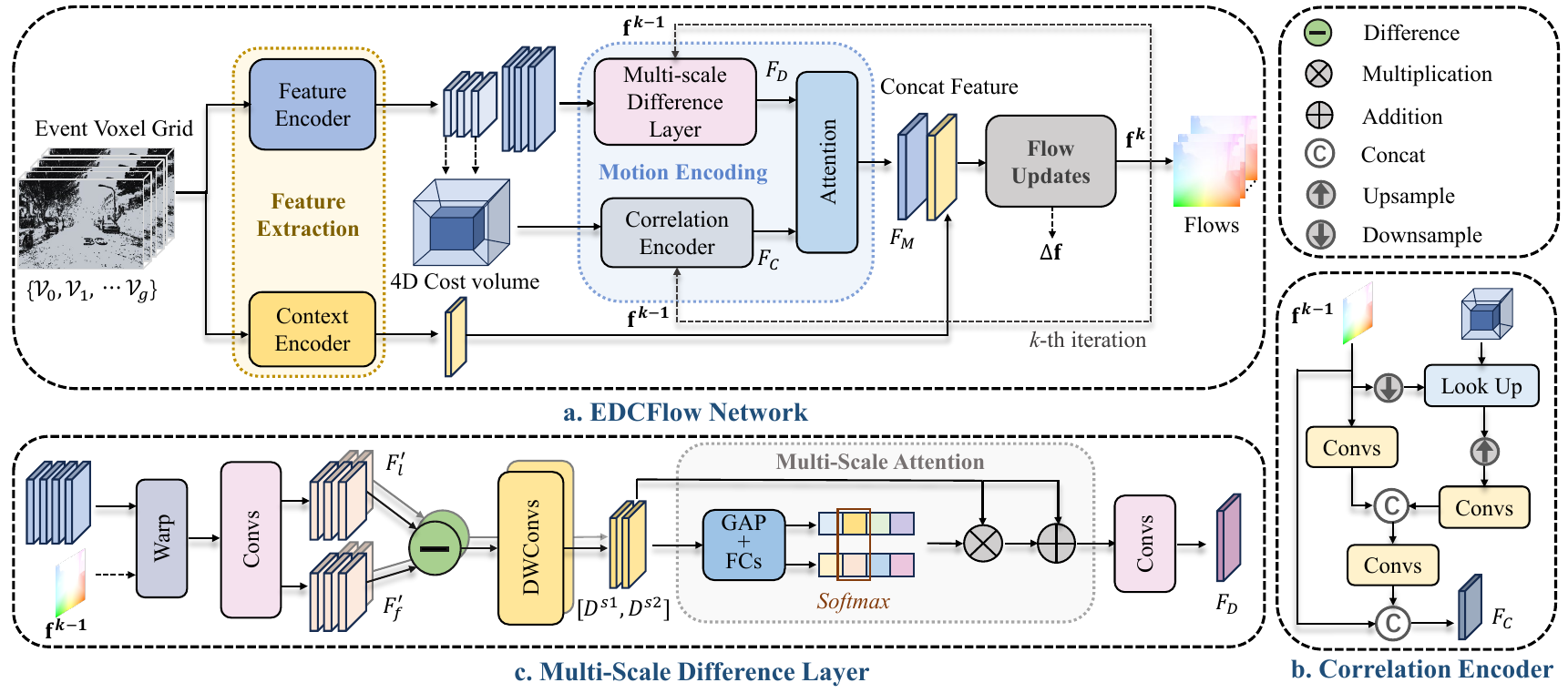}
\caption{\textbf{a}. Overview of \textbf{EDCFlow}, including three main components: 1) \textit{Feature Extraction}. Feature and context encoder extract features from input. 2) \textit{Motion Encoding}. A correlation encoder and multi-scale temporal feature difference layer, along with an attention layer, are utilized to iteratively generate representative motion features. 3) \textit{Flow Updates}. A GRU recurrently updates the flow by decoding the fused motion feature. \textbf{b}. \textbf{Correlation Encoder}. Search correlations at lower resolution, followed by upsampling and high-resolution correlation motion encoding. \textbf{c}. \textbf{Multi-scale Difference Layer}. Compute temporal multi-scale difference motion features at the higher resolution.}
\label{fig_2}
\end{figure*}

\section{Related Work}
\label{sec:Related Work}

\textbf{Optical Flow Estimation on Two-frame.} 
Building on the advancements in deep learning, some U-Net-like methods~\cite{pwc, oiflow, distill, vnc, nlflow} employ similar core components, with a feature pyramid, warping layers, and local correlation reaching high accuracy. Recently, RAFT~\cite{raft} diverges from previous approaches by generating an all-pairs 4D cost volume and using GRU for recurrent flow updates, obtaining high-quality flow. Extended methods~\cite{craft, gma, flowformer} optimize the cost volume to address matching ambiguities. We leverage the excellent framework of RAFT to explore event-based flow estimation.

\noindent\textbf{Event-based Optical Flow Estimation.} 
In event-based vision, early model-based works estimate flow by leveraging the Lucas-Kanade algorithm~\cite{LK-patch, LK}, variational methods~\cite{sofas, pan}, and block-matching~\cite{blockmatch, abmof}. Under the assumption of brightness constancy, some methods~\cite{unifying, secrets} adopt contrast maximization frameworks to improve the flow quality. Most learning-based methods~\cite{EV-FlowNet, ste, zhu_unflow, taming} employ U-Net-like networks to output multi-level flow, but low-resolution outputs lead to details loss. Recently, networks~\cite{e-raft, DCEIFlow, multiflow, TMA, blinkflow} introduce RAFT into the event domain, delivering impressive performance, yet they are inefficient in encoding intermediate motions and face challenges in extending for refinement at higher resolutions. Although IDNet~\cite{idnet} achieves a lightweight network capable of estimating at a higher resolution, it incurs significant computational overhead. In contrast, our approach utilizes dense feature differences to effectively encode continuous motion cues at a higher resolution, achieving better accuracy/model-complexity balance.

\noindent\textbf{Event-based Motion Encoding in Flow Estimation.} 
Some methods~\cite{EST, Matrixlstm, zhu_unflow, EV-FlowNet, EV-MGRFlowNet, ste} tend to use CNNs or RNNs to implicitly learn correspondences. Recent approaches~\cite{blinkflow, multiflow, DCEIFlow, e-raft, TMA} leverage cost volumes to directly construct correspondences, improving the accuracy and robustness. However, they either neglect intermediate motion~\cite{blinkflow, e-raft, DCEIFlow} or adopt computationally expensive multi-cost volume~\cite{TMA, multiflow} to represent it. IDNet~\cite{idnet} employs an RNN backbone to iteratively encode temporal dependencies, yet estimate flow at a higher resolution still requires substantial computations. Instead, we compute feature differences, which effectively capture consistent motion and boundaries~\cite{tea}, aiding intermediate motion encoding in events. However, due to their sensitivity to noise, we propose combining feature differences with a low-resolution cost volume to model representative motion features with greater robustness and efficiency.

%% file: sec/3_method.tex
\section{Method}
\label{sec:EDCFlow}
\subsection{Overall Framework}
\textbf{Event Representation and Problem Description.} Event camera produces an event encoded as $(x, y, t, p)$ by measuring the brightness change, where $(x, y)$, $t$, and $p\in\{-1, +1\}$ denote the coordinate, timestamp, and polarity of the brightness change, respectively. For an event stream $\mathcal{E} = \{(x_{i}, y_{i}, t_{i}, p_{i})\}_{i=0}^{N_{e}-1}$ with $N_{e}$ events, we represent it as a voxel grid with $B$ temporal bins~\cite{zhu_unflow}, with each event contributing its polarity to the two closest temporal bins:
\begin{align}
\mathcal{V}\left(b, x, y\right)&=\sum_{i=0}^{N_e-1} p_{i}\max \left(0,1-\left|b-t_{i}^{*}\right|\right),\label{eq1}\\
t_{i}^{*} &= \frac {(t_{i} - t_{min})}{t_{max} - t_{min}} \times (B - 1),\label{eq2}
\end{align}
where $b\in[0, B-1]$ is the bin index, $t_{max}$ and $t_{min}$ are the maximum and minimum timestamps, respectively. In this work, we aim to estimate the dense flow $\mathbf{f}_{t\to t+1}$ from $t$ to $t+1$ under the two consecutive event sequences $\mathcal{E}_{t-dt\to t}$ and $\mathcal{E}_{t\to t+1}$, with $\mathcal{E}_{t-dt\to t}$ as the reference. 

\noindent\textbf{Overview.} We briefly review RFAT~\cite{raft} in the supplementary material and present an overview of our EDCFlow in~\cref{fig_2} (a). Our EDCFlow consists of three key components: 1) \textit{Feature Extraction.} The feature encoder extracts two-resolution features to encode feature differences at a higher resolution (\textit{i.e.}, 1/4) while constructing a low-resolution (\textit{i.e.}, 1/8) cost volume. The context encoder provides scene context for flow decoding. 2) \textit{Motion Encoding.} At the higher resolution, correlation features are encoded through looking up, motion encoding, and upsampling, while the temporally dense motion features are captured by a multi-scale temporal difference layer. These two complementary motion features are adaptively fused by an attention layer. 3) \textit{Flow Updates.} A GRU decodes the residual flow from the fused motion features and updates the current flow. We iteratively perform motion encoding and flow updates to refine the flow. In the following sections, we will elaborate on the details of our EDCFlow in~\cref{sec:FeatureEX},~\cref{sec:Motion} and~\cref{subsec:loss}.

\subsection{Feature Extraction}
\label{sec:FeatureEX}
The fine-grained temporal detail of event streams provides continuous motion information. To exploit the intermediate motion cues, instead of treating the current event stream $\mathcal{E}_{t\to t+1}$ as a single frame as in E-RAFT~\cite{e-raft}, we divide it into $g$ time windows of size $dt$, along with the reference event stream $\mathcal{E}_{t\to t+1}$, resulting in a total of $g+1$ time windows. Using~\cref{eq1} and~\cref{eq2}, for each time window, we generate event representation $v_i \in \mathbb{R}^{B \times H \times W}$, $i = 0, 1, \dots, g$ , where $H$ and $W$ denote height and width, respectively. We adopt a shared-weight encoder to extract features for $g+1$ voxel grids at 1/4 and 1/8 resolution: $F_i \in \mathbb{R}^{d \times \frac{H}{4} \times \frac{W}{4}}$ and $\bar{F_i} \in \mathbb{R}^{\bar{d} \times \frac{H}{8} \times \frac{W}{8}}$, $i = 0, 1, \dots, g$, where $d$ and $\bar{d}$ represent the channel of features. For saving computation and storage, we build a long-range 4D cost volume $C$ at 1/8 resolution using $\bar{F_0}$ and $\bar{F_g}$:
\begin{equation}\label{eq3}
C = \frac{\bar{F_0} \bar{F_g}}{\sqrt{\bar{d}}} \in \mathbb{R}^{\frac{H}{8} \times \frac{W}{8} \times \frac{H}{8} \times \frac{W}{8}}.
\end{equation}
This cost volume is constructed to provide crucial and robust correspondence information, facilitating the 2D search. The context encoder follows the same architecture as the feature encoder to capture scene contexts for flow updates.

\subsection{Motion Encoding and Flow Updates}
\label{sec:Motion}
We take advantage of high-resolution difference maps and low-resolution cost volume to encode motion features, thus estimating high-quality flow in an efficient manner. As shown in~\cref{fig_2}, we employ an iterative scheme as in RAFT~\cite{raft}, to progressively narrow the flow search space. Each iteration $k\in\{1,2,\cdots, K\}$ involves correlation motion encoding, multi-scale temporal difference motion encoding, attention-based motion fusion, and flow updates.

\noindent\textbf{Correlation Encoder.} 
The difference maps encode changes between adjacent frames but lack direct pixel-level correspondences, whereas the cost volume provides reliable matching cues to capture long-range dependencies. Therefore, integrating correlation motion features from a low resolution offers crucial long-range correspondences efficiently. As shown in~\cref{fig_2} (b), given the estimated flow $\mathbf{f}^{k-1}$ (\textit{i.e.}, $\mathbf{f}^{k-1}_{0\to g}$) at 1/4 resolution, we first downsample it to 1/8 resolution to retrieve correlations from the cost volume $C$. We then upsample the correlation map to 1/4 resolution and follow RAFT~\cite{raft} to encode the correlation motion features $F_C$ through convolutions. 

\noindent\textbf{Multi-scale Temporal Difference Layer.} 
For each iteration, our difference layer strives to capture continuous motion features, providing fine details for inferring the residual flow. We achieve this by warping the target feature maps to the reference one using the current flow before calculating the dense difference maps. Specifically, in~\cref{fig_2} (c), we warp the features $F_{i}$, $i=1,\dots, g$, of consecutive frames, towards $F_{0}$ through bilinear interpolation based on the corresponding flow $\mathbf{f}^{k-1}_{0\to i}$. In small time windows, the motion is minimal, allowing us to assume linear movement. Under this assumption, the flow for each feature map $F_i$ is given by:
\begin{equation}\label{eq4}
\mathbf{f}^{k-1}_{0\to i} = \dfrac{i}{g}\mathbf{f}^{k-1}, i = 1, 2, ..., g.
\end{equation}
 The warped features $\Tilde{F}_{i}$, and the reference feature $F_{0}$, denoted as $\Tilde{F}_{0}$, are considered for motion encoding.

To maintain high computational efficiency, we apply a convolution to reduce the channel dimension of $\Tilde{F}_{i}$, producing former features $\Tilde{F}^{f}_{i} \in \mathbb{R}^{\frac{d}{r} \times \frac{H}{4} \times \frac{W}{4}}, i = 0, 1, ..., g$, where $r$ is the reduction ratio. To alleviate spatial misalignment between adjacent feature maps, we perform channel-wise transformations on the former features to smooth out the spatial boundaries while sharpening the motion boundaries, resulting in the latter features $\Tilde{F}^{l}_{i}$. Considering that fast-moving objects cause large displacements over short durations, while slower movements may exhibit significant changes over longer durations, we introduce a multi-scale sampling strategy to capture multi-scale motion cues. Specifically, we introduce a sampling stride $s$, $s \in \{1, 2, \cdots, g\}$, to control the time interval between the latter and former features, with smaller and higher values capturing fast and slower motions, respectively. A sequence of feature difference maps under a given $s$ is given by:
\begin{equation}\label{eq6}
D_{j}^{s} = \Tilde{F}^{l}_{(j+1)*s} - \Tilde{F}^{f}_{j*s}, j = 0, 1, \dots, \lfloor\frac{g}{s}\rfloor-1.
\end{equation}
We reshape them into $\frac{d}{r} \times \lfloor\frac{g}{s}\rfloor \times \frac{H}{4} \times \frac{W}{4}$. Simply adding these temporal features can lead to the loss of details, while using concatenation or GRU for aggregation introduces significant computational overhead. Therefore, we use two depth-wise separable 3D convolutions (DW-3DConvs) to capture spatio-temporal features while restoring the number of channels, obtaining motion features $D^{s}$. We consider different values of $s = [s1, s2, \cdots]$, and concatenate the produced multi-scale motion features denoted as $D = [D^{s1}; D^{s2};\cdots]$. To accommodate different motion patterns across various scenarios, we design an attention-based multi-scale fusion module, consisting of global pooling, fully connected layers, and softmax along the scale dimension, to adaptively fuse them softly. Finally, a convolution is applied to adjust the channel dimensions and output the difference motion feature $F_D$. This difference motion feature exhibits a stronger ability to handle details at the higher resolution.

\noindent\textbf{Attention-based Motion Fusion.} 
The difference motion features and correlation motion features hold distinct advantages and exhibit complementary properties. To effectively handle the importance variations between these two features in an adaptive manner while maintaining computational efficiency, we employ a simple yet effective channel attention~\cite{hu2018senet} to fuse them:
\begin{equation}\label{eq6}
F_M = {\rm Attention}({\rm Concat}(F_D, F_C)),
\end{equation}
where $F_M$ is the final motion feature for flow estimation. 

\noindent\textbf{Flow Updates.} 
We feed the contextual characteristics and the fused motion features $F_M$ into the GRU unit to output an updated residual flow $\Delta \mathbf{f}$, which is used to calculate the current estimate: $\mathbf{f}^{k} =\mathbf{f}^{k-1} + \Delta \mathbf{f}$.

\subsection{Loss}
\label{subsec:loss}
Following RAFT~\cite{raft}, we supervised our network with $L_{1}$ loss between ground truth $\mathbf{f}^{gt}$ and the predictions $\{\mathbf{f}^{1}, ..., \mathbf{f}^{K}\}$, with exponentially increasing weights, which is given by:
\begin{equation}\label{eqn}
\mathcal{L} = \sum_{k=1}^{K} 0.8^{K-i}\left\|\mathbf{f}^{g t}-\mathbf{f}^{k}\right\|_{1}.
\end{equation}

%% file: sec/4_exper_results.tex
\section{Experiments}
\label{sec:Experiments}
\begin{table*}[t] 
\centering
\resizebox{0.75\linewidth}{!}{
\begin{tabular}{cccccccccc}
    \toprule	
    ~& Method & EPE& AE& 1PE& 2PE& 3PE& Param (M)& MACs (G) & Runt. (ms)\\
		\hline
        MB& MultiCM~\cite{secrets} &  3.47&  13.98&  76.6&  48.4&  30.9& - & - & - \\
        \hline
        USL& TamingCM~\cite{taming} &  2.33&  10.56&  68.3&  33.5&  17.8& - & - & - \\
        \hline
        \multirow{8}{*}{SL}& EV-FlowNet$^{\dagger}$~\cite{EV-FlowNet} &  2.32&  7.9&  55.4&  29.8&  18.6&  14.0&  62 & 7\\
        ~& E-RAFT~\cite{e-raft} &  0.79&  2.85&  12.7&  4.7&  2.7&  5.3&  256 & 102\\
        ~& E-RAFT-4~\cite{e-raft} & \multicolumn{5}{c}{GPU out of memory ($>$ 40GB)}& 5.3&  750 & -\\
        ~& MultiFlow~\cite{multiflow}  &  0.75 & \underline{2.68} &  11.9&  4.4&  2.4& 5.6 & - &-\\
        ~& TMA~\cite{TMA}  & \underline{0.74} & \underline{2.68} &  10.9& 4.0&  2.7&  6.9& 344 & 58\\
        ~& IDNet-8~\cite{idnet}  & 0.77&  3.0&  12.1&  4.0& 2.2&  1.4&  222 & 75\\
        ~& IDNet-4~\cite{idnet}  & \textbf{0.72} & 2.72&  \underline{10.1}&  \textbf{3.5}& \textbf{2.0} & 2.5&  1200 & 120\\
        ~& E-FlowFormer~\cite{blinkflow} &  0.76 & \underline{2.68}&  11.2&  4.1&  2.5& - & - & - \\ 
        ~& \textbf{Ours} & \textbf{0.72} & \textbf{2.65} & \textbf{10.0} & \underline{3.6} & \underline{2.1} & 2.5 & 247 & 39 \\
        \bottomrule
\end{tabular}}
\caption{Evaluation on DSEC~\cite{dsec}. The best results are in bold, while the second-best ones are underlined. $^{\dagger}$ denote the result from E-RAFT~\cite{e-raft}. E-RAFT-4 represents flow estimation at 1/4 resolution.} 
\label{dsec} 
\end{table*}

\begin{figure*}[h]
\centering
\includegraphics[width=0.9\textwidth,height=0.45\textwidth]{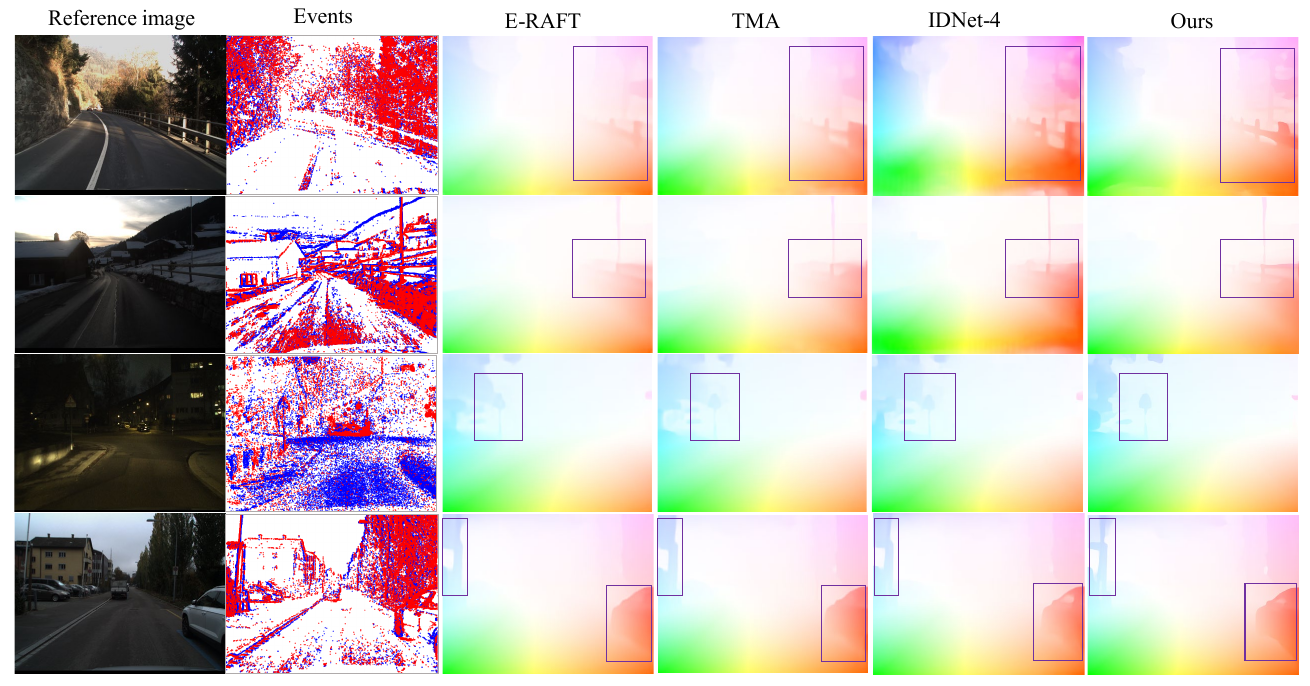}
\caption{\textbf{Qualitative results of flow predictions on DSEC~\cite{dsec}.} Notable areas are marked by bounding boxes. Please zoom in for details.}
\label{dsec_viz}
\end{figure*}

\begin{table*}[h] 
\centering
\resizebox{0.9\linewidth}{!}{
\begin{tabular}{ccccccccccc}
		\toprule	
		\multirow{2}{*}{ }& \multirow{2}{*}{Method}& \multirow{2}{*}{Input}& \multicolumn{2}{c}{\textit{outdoor\_day1}}& \multicolumn{2}{c}{\textit{indoor\_flying1}}&  \multicolumn{2}{c}{\textit{indoor\_flying2}}&  \multicolumn{2}{c}{\textit{indoor\_flying3}}\\
        \cmidrule(lr){4-5}\cmidrule(lr){6-7}\cmidrule(lr){8-9}\cmidrule(lr){10-11}
        ~& ~& ~ & EPE& \% Outlier& EPE& \% Outlier& EPE& \% Outlier& EPE& \% Outlier\\
		\hline
        $dt = 1$& ~ & ~& ~& ~& ~& ~& ~& ~& ~ & ~\\
        \hline
        \multirow{2}{*}{MB}& Brebion $et$ $al.$~\cite{brebion}& E & 0.53&  0.20&  0.52& 0.10&  0.98&  5.50&  0.71& 2.10\\
        ~& MultiCM (Burgers')~\cite{secrets}& E &0.30&  0.10&   \underline{0.42}& 0.10 & \underline{0.60} & \underline{0.59}&  \underline{0.50}& \textbf{0.28}\\
        \hline
        \multirow{2}{*}{USL}& Zhu $et$ $al.$~\cite{zhu_unflow}& E & 0.32& \textbf{0.00}& 0.58& \textbf{0.00}& 1.02& 4.00& 0.87& 3.00\\
        ~& EV-MGRFlowNet~\cite{EV-MGRFlowNet}& E & 0.28&  0.02&  \textbf{0.41}&  0.17& 0.70 &  2.35&  0.59& 1.29\\
        \hline
        \multirow{2}{*}{SSL}& EV-FlowNet~\cite{EV-FlowNet}& E & 0.49& 0.20& 1.03& 2.20& 1.72& 15.1& 1.53& 11.90\\
        ~& STE-FlowNet~\cite{ste}& E & 0.42&  \textbf{0.00}&  0.57&   0.10&  0.79& 1.60&  0.72& 1.30\\
        \hline
        \multirow{5}{*}{SL}& E-RAFT~\cite{e-raft}& E &  0.24&  \textbf{0.00}&  1.10&  5.72&  1.94&  30.79&  1.66& 25.20\\
        ~& DCEIFlow~\cite{DCEIFlow}& E+I$_{1}$ & \textbf{0.22}&  \textbf{0.00}&  0.75&  1.55&  0.90&  2.10&  0.80& 1.77\\
        ~& TMA~\cite{TMA}& E & 0.25&  0.07&  1.06&  3.63&  1.81&  27.29&  1.58& 23.26\\
        ~& IDNet-8~\cite{idnet}& E & 0.33& (\underline{0.01}) & (1.13)  & (2.94) &  (1.67) & (13.64)  & (1.59) & (10.05 )\\
        ~& IDNet-4~\cite{idnet}& E & 0.31& (\textbf{0.00}) & (1.04)  & (1.86) &  (1.55) & (10.84)  &  (1.51) &  (9.58)\\
        \hline
        \multirow{4}{*}{SL}& \textbf{Ours}& E & \underline{0.23} & \textbf{0.00} & 0.97 & 2.19 & 1.49 & 10.79 & 1.38 & 8.98 \\
        ~& \textbf{Ours (+in1)}& E & \underline{0.23} & \textbf{0.00} & - & - & 0.65 & 1.85 & 0.54 & 0.96 \\
        ~& \textbf{Ours (+in2)}& E & \underline{0.23} & \textbf{0.00} & 0.44 & 0.02 & - & - & \textbf{0.49} & \underline{0.34} \\
        ~& \textbf{Ours (+in3)}& E & \underline{0.23} & \textbf{0.00} & \underline{0.42} & \underline{0.01} & \textbf{0.54} & \textbf{0.34} & - & -\\
        \hline
        $dt = 4$& ~ & ~& ~& ~& ~& ~& ~& ~ & ~\\
        \hline
         MB & MultiCM (Burgers')~\cite{secrets}& E & 1.25&  9.21&  1.69&  12.95&  2.49&  26.35&  2.06& 19.03\\
        \hline
        \multirow{2}{*}{USL}& Zhu $et$ $al.$~\cite{zhu_unflow}& E & 1.30& 9.70& 2.18& 24.2& 3.85& 46.80& 3.18& 47.80\\
        ~& EV-MGRFlowNet~\cite{EV-MGRFlowNet}& E & 1.10&  6.22&  1.50&  8.67&  2.39&  23.70&  2.06& 18.0\\
        \hline
        \multirow{2}{*}{SSL}& EV-FlowNet~\cite{EV-FlowNet}& E & 1.23& 7.30& 2.25 & 24.7 & 4.05 & 45.30 & 3.45& 39.70\\
        ~& STE-FlowNet~\cite{ste}& E & 0.99&  3.90&  1.77&  14.70&  2.52&  26.10&  2.23& 22.10\\
        \hline
        \multirow{5}{*}{SL}& E-RAFT~\cite{e-raft}& E & 0.72& 1.12&  2.81&  40.25&  5.09&  64.19&  4.46& 57.11\\
        ~& DCEIFlow~\cite{DCEIFlow}& E+I$_{1}$ & 0.89&  3.19& 2.08& 21.47& 3.48& 42.05& 2.51& 29.73\\
        ~& TMA~\cite{TMA}& E & \underline{0.70}& 1.08&  2.43&  29.91& 4.32&  52.74&  3.60&  42.02\\
        ~& IDNet-8~\cite{idnet}& E &  1.26 &  (4.91)  &  (4.31)  & (58.95) & (6.10)  & (72.97)  & (5.47)  & (68.58) \\
        ~& IDNet-4~\cite{idnet}& E & 1.30 &  (2.48) &  (3.30)  & (42.72) & (4.82)  & (56.82)  &  (4.33) & (52.91) \\
        \hline
        \multirow{4}{*}{SL}& \textbf{Ours}& E &\textbf{0.67} & \underline{0.85} & 2.59  & 28.16  & 4.16 & 44.29 & 3.48 & 36.54\\
        ~& \textbf{Ours (+in1)}& E & \underline{0.70} & 0.89 & - & - & \underline{1.97} & \underline{15.56} & \underline{1.67} & \underline{11.44} \\
        ~& \textbf{Ours (+in2)}& E & \textbf{0.67} & \textbf{0.83} & \textbf{1.28} & \textbf{5.24} & - & - & \textbf{1.47} & \textbf{7.97} \\
        ~& \textbf{Ours (+in3)}& E & 0.72 & 1.29 & \underline{1.30} & \underline{5.56} & \textbf{1.61} & \textbf{11.19} & - & - \\
        \bottomrule
\end{tabular}}
\caption{Evaluation on MVSEC dataset~\cite{mvsec}. E and I$_1$ represent events and reference image, respectively. Results in ( ) are got by re-implementing the source code. `Ours (+in1, 2, 3)' indicates the inclusion of indoor\_flying1, 2, or 3 sequences  as training samples.}
\label{mvsec} 
\end{table*}

\noindent\textbf{Datasets and Metrics.} We conduct experiments on two commonly used datasets: MVSEC~\cite{mvsec}, which has lower resolution and sparser events, resulting in poorer data quality and ground truth, and DSEC~\cite{dsec}, which provides higher resolution and denser events. Following previous works~\cite{e-raft, TMA, idnet}, we train and test our network on the DSEC official training and test sets and report the results on the public benchmark. For MVSEC, we train on two time windows lengths: $dt=1$ (1 grayscale image frame apart) and $dt=4$ (4 grayscale image frames apart), using the outdoor\_day2 sequence for training and testing on indoor\_flying sequences and 800 samples of the outdoor\_day1. We evaluate flow accuracy using end-point-error (EPE) on both datasets. For MVSEC, we additionally measure the percentage of pixels with an EPE greater than 3 pixels and $5 \%$ of ground-truth magnitude (\% Outlier). For DSEC, we compute the angular error in degrees (AE) and percentage of pixels with an EPE greater than n pixels (nPE, n=1, 2, 3). To assess efficiency, we report multiply-accumulate operations (MACs), model size in parameters (Param), and runtime time (Runt.) of per-predicting on the DSEC dataset, using the NVIDIA GeForce RTX 4090.

\noindent\textbf{Implementation Details.} For event representation, we set $g=5$ for both datasets, with $B=3$ for DSEC and MVSEC at $dt=4$, and $B=1$ when $dt=1$, maintaining the same total number of time bins across all time windows as methods~\cite{TMA, e-raft}. We assign downsampling ratio $r=1$ and multiple scales $s=[1, 2, 5]$ for the difference layer. The iteration $K$ and the hidden dimension of GRU are set to 6 and 96, respectively, the channel for correlation motion feature is 64, and the feature extraction follows TMA~\cite{TMA}. Our method, implemented in PyTorch, is trained on DSEC and MVSEC using the AdamW optimizer~\cite{adamw} with a one-cycle learning rate scheduler~\cite{onecycle} (maximum learning rate 0.0002), for 100 and 10 epochs, respectively with a batch size of 3. During training, DSEC and MVSEC are randomly cropped to $288\times384$ and $256\times256$ respectively, with 50\% probability of horizontal flipping and 10\% probability of vertical flipping applied to both datasets. 

\subsection{Results}
\noindent\textbf{DSEC.}~\cref{dsec} presents a quantitative comparison with existing methods on DSEC~\cite{dsec}. Training manners are annotated as MB (model-based), USL (unsupervised learning), and SL (supervised learning). E-FlowFormer is trained on both DSEC and Blinkflow~\cite{blinkflow}, while others use DSEC only. Our method outperforms both model-based and unsupervised approaches in accuracy and surpasses the U-Net-like EV-FlowNet. Among RAFT-like methods~\cite{e-raft, TMA, multiflow, blinkflow}, we achieve the best performance, improving EPE by 9\% over E-RAFT and 3\% over TMA, with better robustness (lower nPE). Compared to IDNet, we perform accuracy comparable to IDNet-4 but with superior AE.

\cref{dsec_viz} illustrates the qualitative comparison of predicted flow, showcasing our predictions against other methods. E-RAFT and TMA estimate flow at the 1/8 resolution, while IDNet-4 and our method estimate directly at the 1/4 resolution. Although TMA aggregates temporally dense motion information to alleviate mismatches in cost volume, performing estimation at low resolution causes a loss of details, leading to lower-quality flow, particularly with blurry motion boundaries. In contrast, our method exhibits clearer details, especially in complex textured scenes, outperforming IDNet-4. This improvement is due to our explicit use of higher-resolution feature differences between adjacent frames, which capture local motion details and boundaries more accurately, whereas IDNet relies on an RNN backbone to implicitly extract pixel correspondences. In addition, the integration of correlation information enables long-range matching, further enhancing our motion representation.

Furthermore, our method demonstrates advantages in model efficiency. It outperforms all RAFT-like models~\cite{e-raft, TMA, multiflow, blinkflow} in terms of model size, computational complexity, and runtime, despite inference flow at twice their resolution. Compared to IDNet-4, which has similar accuracy, our model exhibits only 0.2 times the computation and a 68\% faster runtime due to the repeated RNN backbone computation in IDNet. Although EV-FlowNet achieves faster runtime and IDNet-8 is more lightweight, both sacrifice accuracy. We attribute these advantages to utilizing feature differences rather than multiple cost volumes to capture continuous motions, which not only reduces computational load but also enables flow estimation at higher resolutions. 

\noindent\textbf{MVSEC.}~\cref{mvsec} reports the results on MVSEC~\cite{mvsec} for $dt=1$ and $dt=4$. We annotate the input data types (Input) used for inference in each method. The SSL refers to self-supervised learning methods using photometric error loss from grayscale frames. When trained on the outdoor\_day2 sequence, our method achieves comparable accuracy to DCEIFlow on the outdoor\_day1 at $dt=1$ and SOTA results at $dt=4$, demonstrating its effectiveness in handling both short-term and long-term motion. Similar to E-RAFT, TMA, and IDNet, our method experiences a slight performance drop on other sequences due to the dataset's poor quality and large gap between outdoor and indoor sequences. Despite the challenges posed by the training data, our approach demonstrates the best accuracy on nearly all sequences compared to these methods, highlighting its effectiveness and greater generalizability. The MultiCM and unsupervised methods typically rely on event-based motion compensation, which offers good generalization performance but often entails considerable computational overhead. Moreover, STE and DCEIFlow require the integration of image knowledge. Notably, after incorporating a few indoor samples, our method achieves SOTA performance across all sequences.

\begin{table}[t]
\centering
\resizebox{1.0\linewidth}{!}{
\begin{tabular}{cccccc}
        \toprule	
        Method & Train D.Set & EPE& AE& 1PE& 3PE\\
        \hline
        \multirow{2}{*}{E-RAFT~\cite{e-raft}} & D & 0.80 & 2.76 & 12.5 & 2.9 \\
                                           ~ & \cellcolor{gray!20}B & \cellcolor{gray!20}1.41 (-0.61) & \cellcolor{gray!20}4.76 & \cellcolor{gray!20}39.1 & \cellcolor{gray!20}6.7\\
        \hline
         \multirow{2}{*}{TMA~\cite{TMA}} & D & 0.77 & 2.76 & 11.7 & 2.6\\
                                       ~ &\cellcolor{gray!20} B &\cellcolor{gray!20} 1.49 (-0.72) &\cellcolor{gray!20} 5.73 & \cellcolor{gray!20}42.4 & \cellcolor{gray!20}7.7\\
       \hline
         \multirow{2}{*}{IDNet-4~\cite{TMA}} & D & 0.72 & 2.72 & 10.1 & 2.0 \\
                                           ~ & \cellcolor{gray!20}B &\cellcolor{gray!20} 1.49 (-0.77) &\cellcolor{gray!20} 5.68 &\cellcolor{gray!20}  40.7 &\cellcolor{gray!20} 8.3\\
       \hline
       \multirow{2}{*}{E-FlowFormer~\cite{blinkflow}} & D+B & 0.76 & 2.68 & 11.2 & 2.5\\
                                           ~ &\cellcolor{gray!20} B &\cellcolor{gray!20} 1.33 (-0.57)&\cellcolor{gray!20} \textbf{4.65} &\cellcolor{gray!20} 36.4 &\cellcolor{gray!20} 6.1\\
        \hline
       \multirow{2}{*}{Ours} & D & 0.72 & 2.65 & 10.0 & 2.1 \\
                           ~ &\cellcolor{gray!20} B &\cellcolor{gray!20} \textbf{1.25 (-0.53)} &\cellcolor{gray!20} 4.73 &\cellcolor{gray!20} \textbf{33.3} &\cellcolor{gray!20} \textbf{5.3}\\
        \bottomrule
\end{tabular}}
\caption{Generalization performance on DSEC~\cite{dsec}. D and B refer to the DSEC and Blinkflow~\cite{blinkflow} dataset, respectively. `Reduc.' indicates the percentage reduction in EPE of generalization compared to training directly on DSEC.}
\label{dsec_pretrain} 
\end{table}

\begin{table}[t]
\centering
\resizebox{0.65\linewidth}{!}{
\begin{tabular}{cccccc}
            \toprule
            Model & EPE& AE& 1PE& 2PE& 3PE \\
            \midrule
            W/o Diff & 0.82 & 2.88 & 13.6 & 5.2 & 2.9 \\
            W/o Corr & 0.83 & 3.17 & 14.0 & 5.0 & 2.9\\
            \hline
            W/o SE & 0.74 & 2.69 & 10.4 & 3.7 &  \textbf{2.1}\\
            W/o MSAttn & 0.74 & 2.68 & 10.6 & 3.8 & 2.2\\
            \hline
            \underline{Ours} & \textbf{0.72} & \textbf{2.65} & \textbf{10.0} & \textbf{3.6} & \textbf{2.1} \\
            \bottomrule
\end{tabular}}
\caption{Main components in our model.}
\label{Components} 
\end{table}
\noindent\textbf{Sim-to-real Generalization.} To avoid the influence of low-quality data of MVSEC on the results, we further validate model generalization performance with training on Blinkflow~\cite{blinkflow} and evaluating on DSEC, as shown in~\cref{dsec_pretrain}. We annotate the training dataset (Train D.Set) for each result. Notably, Blinkflow is a simulated dataset. To be fair, we retrain TMA with a batch size of 3 and evaluate E-RAFT without warm-start. The results show that our method achieves the highest accuracy and best robustness compared to other methods. Ours also shows the lowest accuracy drop compared to results directly trained on DSEC, indicating strong generalization to unseen scenes. Furthermore, our model exhibits better sim-to-real generalization, highlighting its transferability and potential to reduce the reliance on large annotated datasets.

\begin{table}[t]
\centering
\resizebox{0.65\linewidth}{!}{
\begin{tabular}{cccccc}
		\toprule	
		Scale & EPE& AE& 1PE& 2PE& 3PE \\
		\hline
        1 & 0.77 &  2.82 & 11.8 & 4.2 &  2.4\\
        2 & 0.74 &  2.72 &  10.7 & 3.9 & 2.2\\
        5 &  0.79 &  2.80 & 12.1  & 4.4 & 2.7\\
        \underline{1, 2, 5} &  \textbf{0.72} & \textbf{2.65}  &  \textbf{10.0} & \textbf{3.6} & \textbf{2.1}\\
        \bottomrule
\end{tabular}}
\caption{Multi-scale strategy.}
\label{Multi-scale} 
\end{table}
\subsection{Ablation Study}
In this section, we provide ablation studies on EDCFlow to show the relative importance of each component. All ablated versions are trained on the DSEC~\cite{dsec} dataset and evaluated on the public benchmark.

\noindent\textbf{Difference Motion Feature.} In~\cref{Components}, we validate the contribution of feature difference (denoted as `W/o Diff', meaning without difference  motion encoding). The results show a 14\% decrease in EPE, indicating that the upsampled correlation at 1/8 resolution is not only coarse but also limited in capturing temporally dense motion information. This also validates that our proposed feature difference strategy between adjacent frames effectively encodes intermediate motion information and better captures local detail.


\begin{figure*}[h]
  \centering
  \begin{subfigure}{0.5\linewidth}
  \centering
    \includegraphics[width=0.7\textwidth,height=0.53\textwidth]{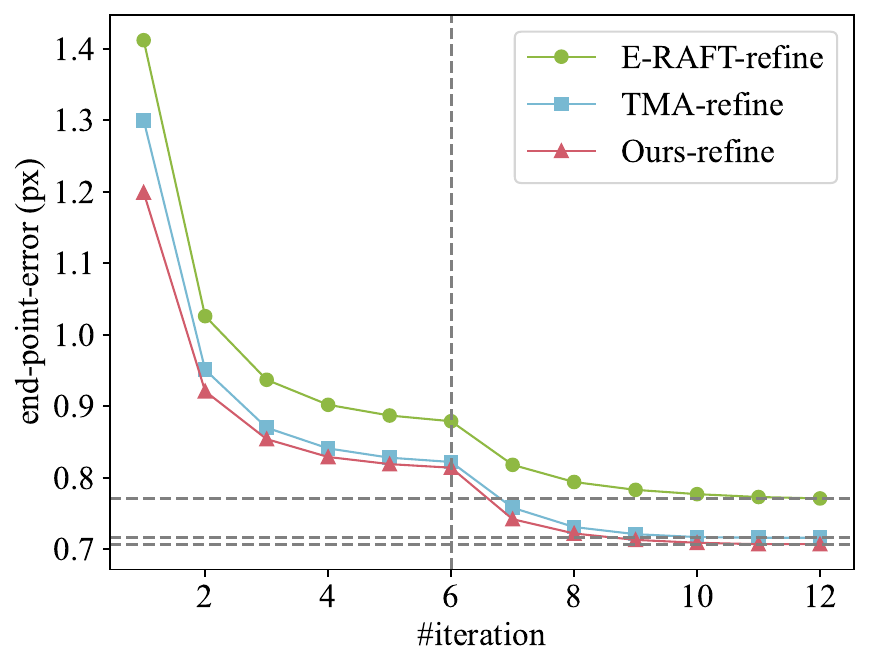}
    \caption{EPE vs. iteration at inference. The addition of refinement stages (starting from iteration=7) results in a notable improvement in accuracy.}
    \label{refine-a}
  \end{subfigure}
  \hfill
  \begin{subfigure}{0.48\linewidth}
  \centering
    \includegraphics[width=0.9\textwidth,height=0.55\textwidth]{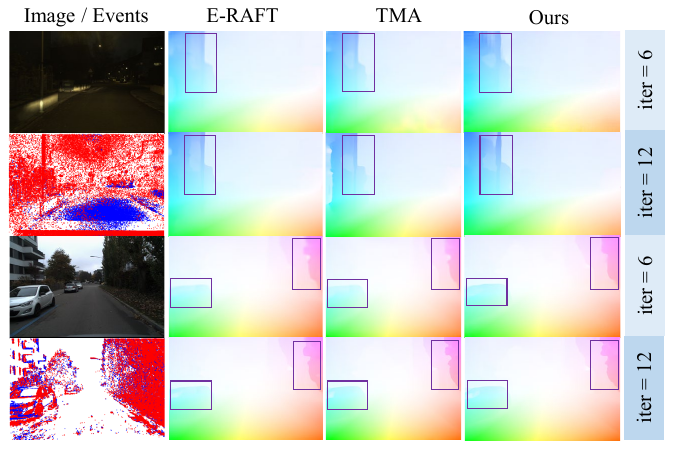}
    \caption{Qualitative results. For each sample, we present the results before refinement (iter=6) and the final flow after refinement (iter=12).}
    \label{refine-b}
  \end{subfigure}
  \caption{Refinement strategy in existing RAFT-like methods.}
  \label{refine}
\end{figure*}

\noindent\textbf{Correlation Motion Feature.} As shown in~\cref{Components}, removing correlation motion encoding (W/o Corr) results in a 15\% accuracy drop. This is because difference maps are sensitive to noise and have lower generalization performance, while correlation offers stronger robustness. Thus, combining both leads to improved accuracy.

\noindent\textbf{Attention Modules.} We study the effects of removing the multi-scale attention mechanism in the feature difference layer (W/o MSAttn) and the channel attention mechanism that fuses correlation and difference motion features (W/o SE). The results in~\cref{Components} indicate that dynamically fusing multi-scale temporal feature difference maps and adaptively aggregating correlation and difference motion features both contribute to further accuracy improvements.

\noindent\textbf{Multi-scale Strategy.} In~\cref{Multi-scale}, we explore the effectiveness of our multi-scale in the difference layer. The scale reflects the sampling interval between adjacent frames. The too-small interval hinders the capturing of subtle motions, and amplifies the changes induced by noise or non-motion, introducing pseudo-motion. Conversely, a large one fails to capture motion details, compromising boundary clarity, particularly in fast-moving or complex scenes. The incorporation of multi-scale concepts can enhance robustness, allowing for adaptation to diverse motion patterns.

\noindent\textbf{Difference Motion Feature Encoding Methods.} We explore various methods to encode spatio-temporal difference motion features within each scale in~\cref{style}. Directly adding consecutive feature difference maps~(\textit{i.e.}, Add.) fails to encode temporal context, limiting the model’s ability to capture continuous motion variations and potentially causing detail degradation or blurring. Both GRU and channel concatenation fused through 2D convolution can capture temporal dependencies, but depthwise separable 3D convolutions provide a more efficient alternative.
\begin{table}[t]
\centering
\resizebox{0.9\linewidth}{!}{
\begin{tabular}{ccccccc}
		\toprule	
		Method & EPE& AE& 1PE& 3PE & MACs (G) & Param (M)\\
		\hline
        Add. & 0.75 & 2.73 & 10.5 & 2.2 & 242  & 2.5 \\
        Concat & \textbf{0.72} & \textbf{2.65} & \textbf{9.8} & 2.1 &  311 &  3.1 \\
        GRU &  0.73 & 2.69 &  9.9 &  \textbf{2.0} & 445 &  2.7 \\
        \underline{DWConv3d} & \textbf{0.72} & \textbf{2.65} & 10.0 & 2.1 & 247 & 2.5 \\
        \bottomrule
\end{tabular}}
\caption{Difference motion feature encoding methods.}
\label{style} 
\end{table}

\subsection{Extending to RAFT-like methods}
We integrate the proposed EDCFlow after existing RAFT-like networks as a refinement stage to obtain higher-quality flow. We construct a 1/8 resolution version of EDCFlow (\textit{i.e.}, Ours-8). `Methods-refine' represents running the original flow computation process with 6 iterations at 1/8 resolution, followed by 6 iterations of our refinement at 1/4 resolution with the hidden dimension of 64 for GRU. For comparison, we evaluate all methods with results from 6 and 12 iterations (denoted as $\times$6 and $\times$12) at the 1/8 resolution. The results in~\cref{higher-refinement} demonstrate that merely increasing iteration count at 1/8 resolutions offers limited accuracy gains, as the estimation converges after several iterations, leaving the flow near a local optimum. Conversely, refinement at higher resolutions leads to substantial accuracy improvements without significantly increasing the model size.~\cref{refine-a} intuitively provides a depiction of the accuracy trajectory when incorporating higher-resolution refinement, revealing a notable error reduction at iteration=7. Qualitative results in~\cref{refine-b} further highlight that refined flow enhances details, especially along motion boundaries and in regions with intricate textures.

\begin{table}[t]
\centering
\resizebox{1.0\linewidth}{!}{
\begin{tabular}{ccccccc}
		\toprule	
		Method & EPE& AE& 1PE& 3PE & MACs (G) & Param (M)\\
		\hline
        E-RAFT\_$\times$6~\cite{e-raft} & 0.82 & 2.83 & 13.1  & 3.0 & 161 & 5.3\\
        E-RAFT\_$\times$12~\cite{e-raft} & 0.80 & 2.76 & 12.5 & 2.9 & 256 &5.3\\
        E-RAFT-refine~\cite{e-raft} & 0.77 &  2.81 & 11.1 & 2.5 & 364 & 6.16\\
        \hline
        TMA\_$\times$6~\cite{TMA} & 0.77 & 2.76 &  11.7 & 2.6 & 344 & 6.9\\
        TMA\_$\times$12~\cite{TMA} & 0.76 & \textbf{2.66} &  11.2 & 2.4 & 620 & 6.9\\
        TMA-refine~\cite{TMA} & 0.72 & \textbf{2.66} & \textbf{9.2} & 2.0 & 478 & 7.9 \\
        \hline
        Ours-8\_$\times$6 & 0.78 & 2.82 & 12.7 & 2.5 & 216 & 6.8\\
        Ours-8\_$\times$12 & 0.79 & 2.89 & 13.89 & 2.6 & 367 & 6.8\\
        Ours-8-refine & \textbf{0.71} & 2.72 & 9.8 & \textbf{1.9} & 352 & 8.0\\
        \bottomrule
\end{tabular}}
\caption{Higher-resolution refinement in RAFT-like methods.}
\label{higher-refinement} 
\end{table}


%% file: sec/5_conclusion.tex
\section{Conclusion}
\label{sec:Conclusion}
In this paper, we explore the multi-scale temporally dense feature differences to efficiently capture continuous motions. Leveraging the complementarity between these features and cost volume, we construct EDCFlow, a lightweight network for event-based optical flow estimation. By adaptively fusing difference motion features with relatively low-resolution correlations, EDCFlow enables high-quality flow inference at a higher resolution. Comprehensive experiments on DSEC and MVSEC validate the superior performance and efficiency of our method, achieving a balance between model accuracy and complexity. Particularly, our method can be flexibly integrated into RAFT-like networks for further refinement.

\noindent\textbf{Acknowledgment.}
This work was supported by the National Natural Science Foundation of China (Grant No. 62273093, 62236002).

%% file: sec/X_suppl.tex
\clearpage
\setcounter{page}{1}
\maketitlesupplementary

\section{More comparison with SOTA methods}
\noindent\textbf{Compared with frame-based methods.} Several frame-based SOTA methods~\cite{dip, gmflow+, ccmr} focus on achieving high-resolution optical flow estimation. They primarily perform pixel matching or refinement across multiple spatial resolutions in frames. Unlike frames, event data provides high temporal resolution, necessitating efficient method to capture continuous motion features. In this context, our multi-scale difference layer effectively handles temporal dynamics by leveraging temporal feature differences with low computations and can refine RAFT-like networks. Using publicly available source code under the same training settings, our method outperforms these methods in accuracy and efficiency on DSEC (in~\cref{framecomparison}), highlighting its importance in event-based flow estimation.
\begin{table*}[h] 
\centering
\resizebox{0.7\linewidth}{!}{
\begin{tabular}{ccccccccc}
    \toprule	
    Method & EPE& AE& 1PE& 2PE& 3PE& Param (M)& MACs (G) & Runt. (ms)\\
		\hline
         DIP~\cite{dip} &  3.06&  10.95&  77.8&  46.6&  27.5& 5.4 & 590 & 92 \\
        GMFlow+~\cite{gmflow+} &  6.55&  7.40&  94.1& 77.9&  63.6& 4.6 & 223 & 141 \\
        CCMR~\cite{ccmr} & \multicolumn{5}{c}{GPU out of memory ($>$ 40GB)}& 11.6 & 2255 & -\\
        \textbf{Ours} & \textbf{0.72} & \textbf{2.65} & \textbf{10.0} & \textbf{3.6} & \textbf{2.1} & 2.5 & 247 & 39 \\
        \bottomrule
\end{tabular}} \
\caption{Compared with frame-based SOTA methods.}
\label{framecomparison}
\end{table*}

\begin{figure}[h]
  \centering
  \includegraphics[width=0.65\linewidth, height=0.6\linewidth]{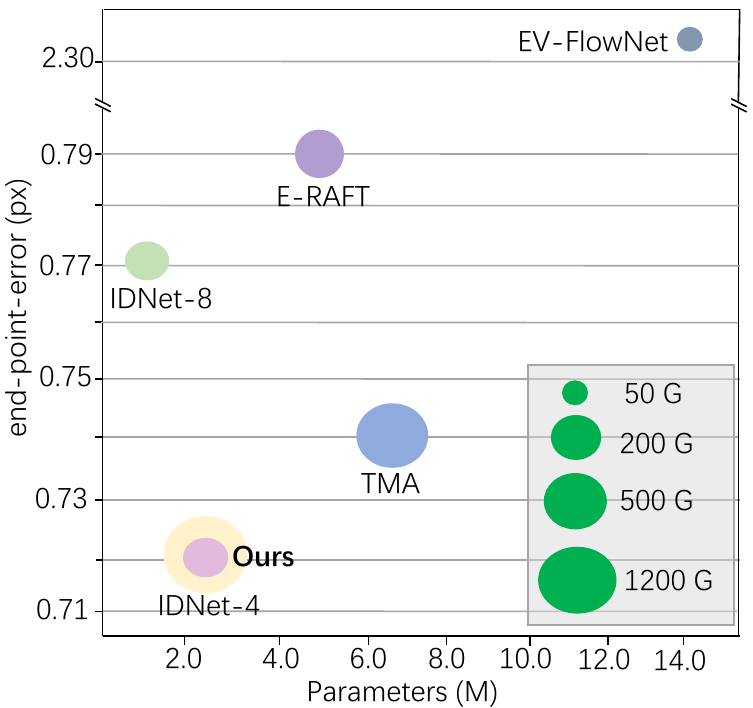}
\caption{End-point-error (px) on DSEC vs. computational complexity (MACs: G) and model size (Parameters: M). All models are trained on DSEC, and tested with one NVIDIA 4090 GPU. The computational complexity  corresponding to the size of the circle is shown in the legend in the lower right corner.}
\label{qipaotu}
\end{figure}

\noindent\textbf{Compared with event-based methods.}
We present accuracy vs. complexity in~\cref{qipaotu} to facilitate comparison. Our EDCFlow achieves higher performance as well as significant reductions in computational overhead over the state-of-the-art methods.

\section{More Visualizations}
\begin{figure*}[t!]
\centering
\includegraphics[width=1.0\textwidth,height=0.45\textwidth]{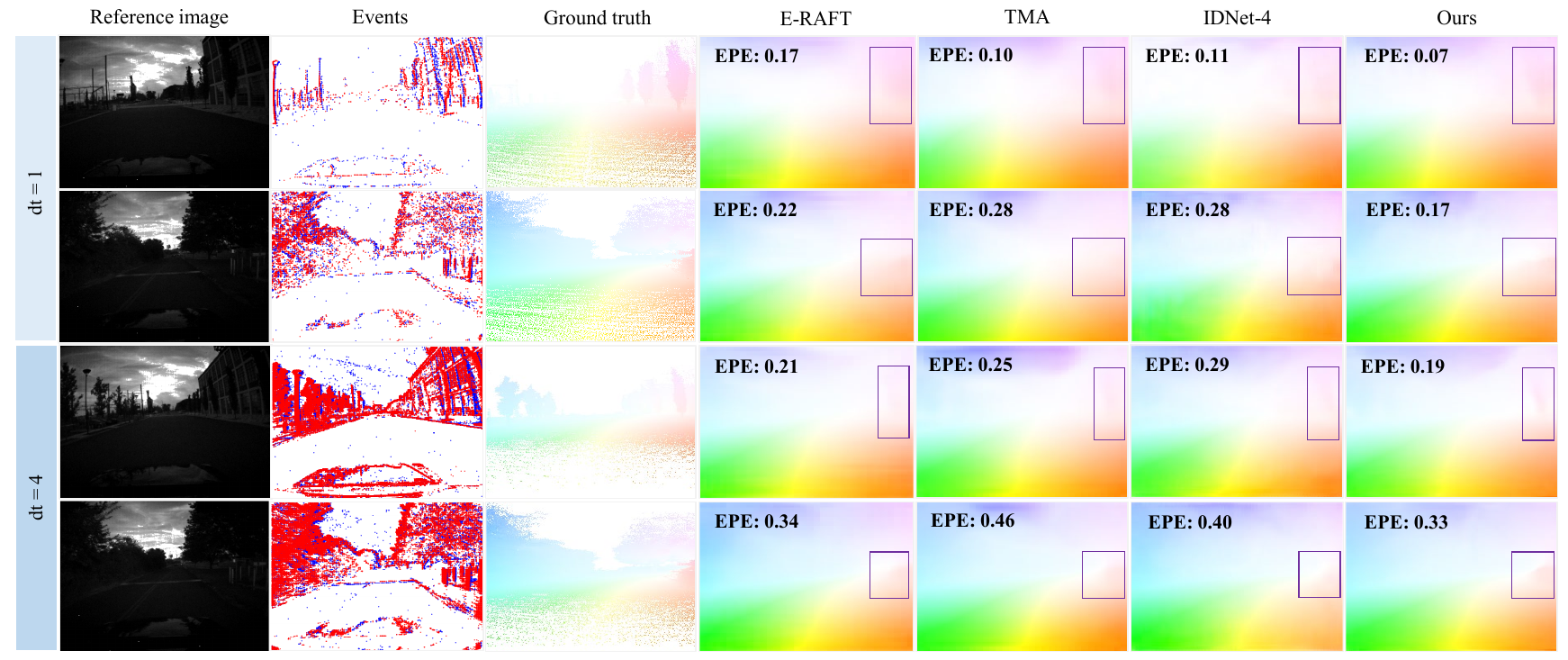}
\caption{Qualitative results on the outdoor\_day1 sequence on MVSEC~\cite{mvsec}. Please zoom in for details.}
\label{mvsec_viz}
\end{figure*}

\noindent\textbf{Qualitative Results on MVSEC.}~\cref{mvsec_viz} presents a qualitative comparison of our method with other methods on outdoor\_day1 sequence of the MVSEC~\cite{mvsec}. Compared to DSEC dataset~\cite{dsec}, MVSEC has lower resolution and sparser events (especially at $dt=1$), and it lacks occlusion and moving objects handling for its ground truth, making the data quality relatively poor. Despite this, compared to other methods, our approach holds superior performance at both $dt=1$ and $dt=4$, capturing clearer motion boundaries. This confirms that our method demonstrates greater adaptability across diverse data distributions and scenes.

\begin{figure}[t]
\centering
\includegraphics[width=0.5\textwidth,height=0.52\textwidth]{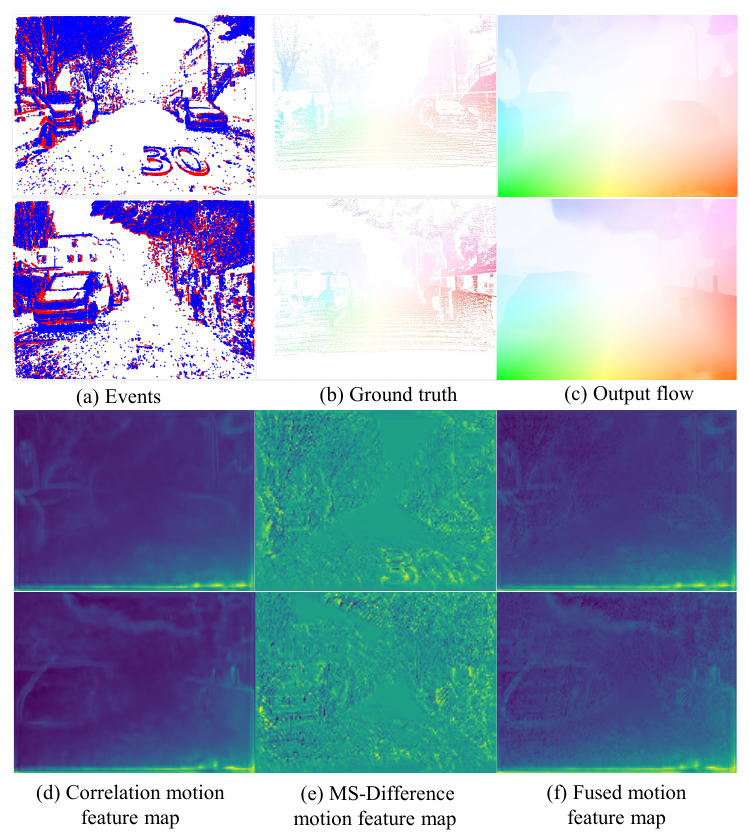}
\caption{Illustration of motion feature maps. (a) Event data. (b) Optical flow ground truth. (c) Optical flow estimated by our method. (d) Correlation motion feature map. (e) Multi-scale temporal difference motion feature map fused with multi-scale attention. (f) The final motion feature map is aggregated from the difference motion feature and correlation motion feature.}
\label{motionfeat}
\end{figure}

\noindent\textbf{Motion Feature Maps.} To better illustrate the complementarity between the difference maps and correlation maps, we present in~\cref{motionfeat} the multi-scale temporal feature differences of adjacent feature maps and the motion features encoded from the cost volume on the DSEC dataset. The difference motion features exhibit strong responses in textured areas but are noisy, while the correlation motion features may produce blurred boundaries due to matching ambiguities. By adaptively fusing these two features, the response at motion boundaries can be enhanced, resulting in high-quality optical flow.

\begin{figure*}[t]
\centering
\includegraphics[width=1.0\textwidth,height=0.6\textwidth]{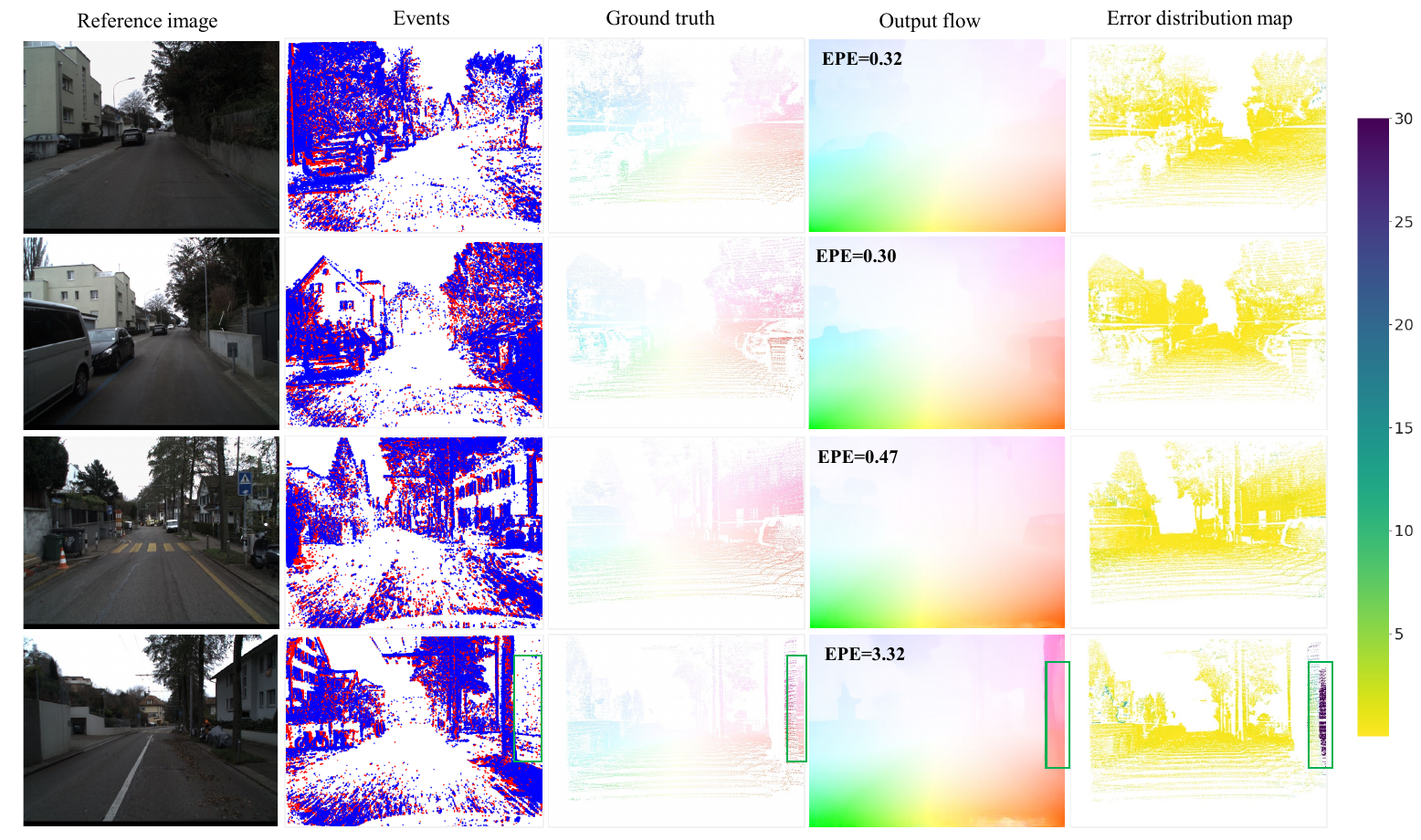}
\caption{Visualization of error distribution maps. We present three high-quality flow estimation results with smaller EPE (the first three columns) and one failure case with larger EPE (the last column). The color bar indicates the per-pixel square root error magnitude, with darker colors representing larger errors.}
\label{errormap}
\end{figure*}

\noindent\textbf{Flow Error Maps.} To analyze the advantages and limitations of the model, we visualize the EPE distributions in~\cref{errormap}, where the error for each pixel is represented as the square root error between the estimated and ground truth flows, shown using heatmaps. Since the ground truth of the DSEC dataset's test set is not publicly available, we use zurich\_city\_05\_b and zurich\_city\_11\_c from the training set as the test set for error analysis, while the remaining training samples are used to train the model. Our model can estimate accurate flow in most scenarios, particularly for complex textured objects like trees. In failure cases, however, our model encounters significant estimation errors in the spatial edge regions. This may be due to two reasons: first, the sparse events in these regions lack sufficient texture information, making it difficult for feature differences to encode motion boundaries and for the cost volume to resolve matching ambiguities; second, pixels at the spatial edges cannot aggregate enough contextual information to encode accurate motion features. These issues could be addressed by fusing images and leveraging multiple preceding and subsequent event streams.

\section{More Ablation Studies}
More ablation studies are also conducted on the DSEC dataset~\cite{dsec}.

\noindent\textbf{Iterations.} Some existing methods~\cite{ste, raft, idnet} achieve better optical flow results through iterative refinement strategies, particularly for small objects with large displacements. The results in~\cref{iterations} show that as the number of iterations increases, the flow results become stable and reach a convergent state. The performance stabilizes when the number of iterations reaches 6. Moreover, excessive iterations may cause overfitting or oscillation around a local optimum, such as ours-8\_6/12 in~\cref{higher-refinement}.

\begin{table}[h]
\centering
\resizebox{0.6\linewidth}{!}{
\begin{tabular}{cccccc}
		\toprule	
		iterations & EPE& AE& 1PE& 2PE& 3PE \\
		\hline
        2 & 0.86 & 3.11 & 13.8 & 5.3 & 3.1\\
        4 & 0.77 &  2.85 & 11.5 & 4.2 & 2.4\\
        \underline{6} &  \textbf{0.72} & 2.65  &  \textbf{10.0} & \textbf{3.6} & \textbf{2.1}\\
        8 & \textbf{0.72} & \textbf{2.64} &  10.1 & \textbf{3.6} & \textbf{2.1}\\
        \bottomrule
\end{tabular}}
\caption{Iterations. The number of iterative refinements.}
\label{iterations} 
\end{table}

\noindent\textbf{Event splitting.} The time windows and temporal bins determine the input's temporal resolution and the number of input channels, respectively. The results in~\cref{split} show that a small value of $g$ leads to performance degradation due to the loss of intermediate motion information. When the $g$ is set to 5 and $B$ to 3, sufficient continuous motion features are captured, achieving good performance. Increasing the window count further achieves comparable performance but introduces additional computational overhead.

\begin{table}[h]
\centering
\resizebox{0.9\linewidth}{!}{
\begin{tabular}{cccccccc}
        \toprule	
        $g$, $B$ & EPE& AE& 1PE& 2PE& 3PE & Param (M) & MACs (G)\\
        \hline
        1, 15 & 0.79 & 2.80 & 11.9 & 4.4 & 2.9 & 2.2 & 182\\
        3, 5 & 0.75 & 2.74 & 10.7 & 3.9 & 2.3 & 2.3 &209\\
        \underline{5}, \underline{3} &  \textbf{0.72} & \textbf{2.65}  &  \textbf{10.0} & 3.6 & 2.1 & 2.5 & 247\\
        15, 1 & \textbf{0.72} & 2.68 &  \textbf{10.0} & \textbf{3.5} & \textbf{2.0} & 2.5 & 322\\
        \bottomrule
\end{tabular}}
\caption{Event splitting. $g$ represents the number of time windows. $B$ denotes the number of time bins.}
\label{split} 
\end{table}

\noindent\textbf{Optical flow estimation resolution.} Different resolutions capture varying degrees of detail, offering distinct levels of granularity in the information provided. As shown in~\cref{resolution}, flow estimation performs best at 1/4 resolution. We attribute this to the fact that the feature difference strategy is more effective in capturing local detail, making it more accurate at higher resolutions, while at 1/8 resolution, some detail is inevitably lost. Although the 1/2 resolution retains more detail, its tolerance to noise decreases, and the correlation features upsampled by two times remain too coarse to robustly enhance the final motion feature representation. Additionally, this resolution struggles to handle large displacements effectively.

\begin{table}[h]
\centering
\resizebox{0.9\linewidth}{!}{
\begin{tabular}{ccccccc}
		\toprule	
		Resolution & EPE& AE& 1PE& 3PE & MACs (G) & Param (M)\\
		\hline
        8 & 0.78 & 2.82 & 12.7 & 2.5 & 216 & 6.8 \\
        \underline{4} & \textbf{0.72} & \textbf{2.65} & \textbf{10.0} & \textbf{2.1} & 247 & 2.5 \\
        2 & 0.77 & 2.89 & 10.8 & 2.5 & 288 & 1.1 \\
        \bottomrule
\end{tabular}}
\caption{Optical flow estimation resolution. The resolutions of 8, 4, and 2 represent flow computed at 1/8, 1/4, and 1/2 resolution of the input, respectively.}
\label{resolution} 
\end{table}

\begin{table}[h]
\centering
\resizebox{0.9\linewidth}{!}{
\begin{tabular}{ccccccc}
		\toprule	
		$r$ & EPE & AE& 1PE& 3PE & MACs (G) & Param (M)\\
		\hline
        \underline{1} & \textbf{0.72} & \textbf{2.65} & \textbf{10.0} & \textbf{2.1} & 247 & 2.5 \\
        2 & 0.74 & 2.65 & 10.4 & 2.2 & 244 & 2.5 \\
        8 & 0.75 & 2.73 & 11.0 & 2.2 & 241 & 2.5 \\
        w/o & 0.73 & 2.67 & 10.2 & 2.2 & 244 & 2.5 \\
        \bottomrule
\end{tabular}}
\caption{Reduction ratio in the multi-scale temporal difference layer.}
\label{r} 
\end{table}

\noindent\textbf{Feature dimension reduction ratio.} In~\cref{r}, we explore the impact of $r$ in the multi-scale temporal difference layer, with ``w/o'' indicating conv1 removed. We set $r=1$ to balance accuracy and computations. When compared to ``w/o'', setting $r=1$ brings imporvements in accuracy by leveraging conv1 to enhance feature interaction. Since the channel number is 64 in our experiment, the effect of $r$ on computation is marginal. For larger channel numbers, $r>1$ can be  used to balance accuracy and computation.

\section{Future Work} 
In our work, we assume linear motion within short time windows (20 ms for DSEC and 10/40 ms for MVSEC), which shows good empirical performance and lower computational complexity. However, investigating alternative motion models, such as estimating higher temporal resolution intermediate flows to capture complex trajectories, could be an interesting direction for future research. Furthermore, event-image fusion-based optical flow estimation represents a promising research direction. By effectively combining the high temporal resolution and motion-capturing capability of event streams with the rich appearance and texture information provided by images, the accuracy and robustness of optical flow estimation can be significantly enhanced.